\documentclass[11pt]{article}
\usepackage[margin=1in]{geometry}
\usepackage{amsmath,amssymb,amsthm}
\usepackage{algorithm}
\usepackage{algpseudocode}
\usepackage{booktabs}
\usepackage{graphicx}
\usepackage{float}
\usepackage{tikz}
\usetikzlibrary{arrows.meta,calc}
\usepackage{hyperref}
\hypersetup{colorlinks=true, linkcolor=blue, citecolor=blue, urlcolor=blue}
\usepackage{microtype}
\usepackage{xcolor}
\usepackage{siunitx}
\usepackage{natbib}
\DeclareFontShape{OT1}{cmr}{bx}{sc}{<->ssub*cmr/bx/n}{}
\title{NoHairCache: Calibrate-Once KV Cache Eviction with One Power-Law Exponent}
\author{Shuyang Xiang}
\date{}

\newtheorem{proposition}{Proposition}
\newtheorem{theorem}{Theorem}
\newtheorem*{remark}{Remark}

\begin{document}
\maketitle

\begin{abstract}
The decay exponent $\alpha$ in KV-cache eviction is an intrinsic property of the model: calibrated once on a single corpus, it transfers directly across domains. Attention-based accumulation (H2O) and recency-based sliding windows (StreamingLLM) are two dominant eviction frameworks, often treated as unrelated, but we show that they are the two endpoints of a single power-law family: by scoring each token as $s_i = A_i \cdot (t-i)^{-\alpha}$, where $A_i$ is cumulative attention and $\alpha$ controls the age penalty, H2O and StreamingLLM correspond to $\alpha=0$ and $\alpha\to\infty$, respectively (with StreamingLLM additionally hard-coding an attention-sink patch).

The no-hair mechanism of \citet{xiang2026nohair} bounds the admissible range of $\alpha$, and the measured exponents agree: calibrating once on WikiText yields $[0.556, 0.822]$ for a representative model (Qwen2.5-1.5B-Instruct). The per-model ranges are consistent across five models (410M--1.5B) and independently match the optimal regions from perplexity sweeps.

\textsc{NoHairCache} achieves the lowest perplexity across three datasets (WikiText, PG19, and arXiv) without per-dataset tuning, outperforming both H2O and StreamingLLM ($p < 0.05$ on all three datasets, with $0.06$--$0.10$ PPL gains over StreamingLLM at budget $512$). On downstream four-way multiple-choice QA, it matches or outperforms StreamingLLM across all nine (task, budget) cells, retaining $98.2\%$ of full-cache accuracy on NarrativeQA at $25\%$ retention versus $91.1\%$ for StreamingLLM; per-cell McNemar tests are not significant, so QA is corroborating only. We release our calibration script and drop-in scorer upon publication.
\end{abstract}

\section{Introduction}

Long-context inference is often bottlenecked at generation time: every decoded token has to attend to the entire KV cache, causing memory usage to grow linearly and, as a result, total attention compute to scale quadratically with sequence length. Caching every past key-value pair is therefore untenable, making token eviction unavoidable. The question now is: \emph{by what criterion should tokens be evicted?}

Existing eviction policies rely on diverse scoring heuristics: accumulated attention (H2O) \citep{zhang2023h2o}, recency windows (StreamingLLM) \citep{xiao2024streaming}, observation-window pooling (SnapKV) \citep{li2024snapkv}, query-aware relevance (Quest) \citep{ge2024model}, or even random selection. Two paradigms dominate in practice: accumulation-based methods (e.g., H2O) and recency-based methods (e.g., StreamingLLM) \citep{zhang2023h2o,xiao2024streaming}. Although power-law decay of token influence has been observed empirically \citep{lin2017critical,xiang2026nohair}, these two paradigms remain disconnected heuristics without a unified parameterization. 

In this work, we show that they are in fact opposite endpoints of a single power-law family. By scoring each cached token $i$ at decoding step $t$ as
\begin{equation}
s_i = A_i \cdot (t-i)^{-\alpha},
\end{equation}
where $A_i$ denotes cumulative attention and $\alpha \ge 0$ governs the age penalty, we are able to unify both strategies under a single framework. The H2O and StreamingLLM scoring rules then correspond to the endpoints $\alpha=0$ and $\alpha\to\infty$, respectively (StreamingLLM additionally hard-codes an attention sink). Consequently, designing an eviction policy reduces from choosing between competing heuristics to identifying the decay exponent $\alpha$ of this family, which we calibrate per layer.

The exponent is an intrinsic property of the model architecture rather than the input data distribution. Measuring $\alpha$ for each model once on a single calibration corpus (e.g., WikiText-103) yields consistent means and bulk ranges across models (such as Qwen2.5-1.5B and Pythia-1.4B), although the detailed layer-wise profiles are only partly preserved; the measured values align closely with optimal perplexity sweeps, and each model's frozen $\alpha$ transfers unchanged to distinct domains like PG19 and arXiv.

To ground $\alpha$ theoretically, we draw on the empirical \emph{no-hair mechanism} established in our prior work \citep{xiang2026nohair}, which shows that a prompt's hidden-state influence on position $t$ decays as $t^{-\alpha}$, with $\alpha$ bounded layer-wise within $[0.2, 0.7]$ by attention-dilution geometry. The theorem constrains the exponent within a family but does not select the family itself.

Building on this insight, we introduce \textsc{NoHairCache}, a lightweight, drop-in scorer that replaces the scoring function of top-$k$ eviction policies with our power-law formulation based on offline-calibrated, per-layer $\alpha$ values, incurring only a single scalar multiplication per eviction step over H2O. In summary, we make the main following contributions:
\begin{itemize}
  \item \textbf{Intrinsic, Calibrate-Once Decay Profiles.} We demonstrate that the per-layer decay exponent $\alpha$ is an intrinsic property of the model rather than the corpus: across five models its mean lies in $0.602$--$0.663$, and a single offline calibration transfers without re-tuning across WikiText, PG19, arXiv, cache budgets, and downstream QA. We further show why hidden-state influence avoids the softmax recency bias that underestimates $\alpha$ in attention-mass proxies.
  \item \textbf{Unified Power-Law Framework.} We relate accumulation-based (H2O) and recency-based (StreamingLLM) KV-cache eviction as opposite endpoints ($\alpha=0$ and $\alpha\to\infty$) of a single power-law decay family $s_i = A_i \cdot (t-i)^{-\alpha}$; the family is a modeling choice, not a consequence of the no-hair theorem, which only constrains the exponent within a family.
  \item \textbf{Lightweight Drop-In Algorithm (\textsc{NoHairCache}).} We propose a training-free, plug-and-play eviction scorer that integrates per-layer calibrated exponents into standard top-$k$ eviction pipelines with negligible overhead (a single scalar multiplication per step over H2O).
  \item \textbf{Cross-Domain and Downstream Gains.} Across three corpora (WikiText, PG19, arXiv) and three long-context QA benchmarks (NarrativeQA, HotpotQA, Qasper), \textsc{NoHairCache} improves perplexity with statistical significance (paired Wilcoxon $p<0.05$ on every dataset), while on QA it is never worse than StreamingLLM (e.g., retaining $98.2\%$ of full-cache accuracy on NarrativeQA at a $25\%$ cache budget vs.\ $91.1\%$ for StreamingLLM) though the gains are not statistically significant (per-cell McNemar).
\end{itemize}

Echoing the ``no-hair'' principle in black hole physics, which argues that complex objects collapse down to just a few fundamental parameters, our findings suggest that the seemingly complex behavior of KV cache eviction policies is largely governed by a single intrinsic power-law form, whose exponent profile is fixed once per model. Rather than designing increasingly intricate heuristics, future cache compression methods can build directly on this unified power-law formulation with a one-time, per-model calibration.

\paragraph{Organization.} We organize this work as follows. 
Sec.~\ref{sec:related} reviews prior work on long-context degradation, attention mechanics, and KV-cache compression. Sec.~\ref{sec:kv_eviction} introduces the unified score family and its limiting cases, gives the power-law form, and presents the calibration-and-eviction algorithm. Sec.~\ref{sec:experiments} reports the measured exponents across models and scales, confirms them against perplexity sweeps, and evaluates cross-dataset transfer, budget regimes, cross-model robustness, and downstream QA. Finally, Sec.~\ref{sec:discussion} discusses limitations, and Sec.~\ref{sec:conclusion} concludes.

\section{Related Work}
\label{sec:related}

Our work bridges theoretical analyses of information propagation in Transformers with practical memory management for long-context inference. We build our contributions across three primary lines of research.

\paragraph{Long-Context Degradation and Instruction Drift.}
A well-documented failure mode of autoregressive language models is the gradual erosion of early context and system instructions over extended generation horizons \citep{li2024instability,liu2024lost}. Even when prompts remain within the nominal context window, models exhibit recency bias and fail to retrieve or adhere to initial constraints. To stabilize infinite-length generation, \citet{xiao2024streaming} identified the phenomenon of \emph{attention sinks}, which is, the disproportionate allocation of attention mass to initial tokens, and proposed retaining them alongside a local sliding window. Subsequent theoretical work by \citet{barbero2025first} suggests that these sinks act as a mechanism to prevent over-mixing across layers. Training-dynamics measurements in \citet{xiang2026nohair} show that prompt forgetting sets in much earlier in training (roughly 0.3--1B tokens) than sink formation, so sinks are neither the initial cause of forgetting nor a reliable proxy for prompt retention.

\paragraph{Attention Mechanics, Over-Squashing, and Information Flow.}
Understanding how information travels across tokens and layers has been extensively studied through norm-based decompositions \citep{kobayashi2020norm} and attention flow graphs \citep{abnar2020flow}. \citet{dong2021rank} showed that pure self-attention suffers from doubly exponential rank collapse with depth, while \citet{noci2022signal} analyzed signal propagation at initialization using mean-field theory. Along the sequence axis, \citet{barbero2024glasses} formalized \emph{over-squashing} in decoder-only Transformers, bounding the sensitivity of final representations to distant tokens. Complementing this, \citet{nakanishi2025ssmax} observed that softmax distributions flatten as context length grows (``attention fading'') and proposed length-dependent logit scaling (Scalable-Softmax). Furthermore, natural language itself exhibits power-law decay in long-range mutual information \citep{lin2017critical}. The conditional no-hair theorem of \citet{xiang2026nohair} provides an asymptotic bound of $O(P(\log t)^{L-1}/t)$ on prompt offset decay, isolating the exact roles of direct attention dilution and multi-layer token relay. We take its finite power-law exponent as the quantity to calibrate here.

\paragraph{KV Cache Eviction and Compression.}
To mitigate the memory bottleneck of long contexts, numerous heuristics have been proposed to evict unimportant tokens from the KV cache during inference. Heavy-Hitter Oracle (H2O) \citep{zhang2023h2o} greedily retains tokens with the highest cumulative attention scores, exploiting the sparsity of attention weights. Similarly, ScissorHands \citep{liu2023scissorhands} leverages the persistence of pivotal tokens across decoding steps, while SnapKV \citep{li2024snapkv} selects clustered important KV positions based on observation windows. Other approaches allocate cache budgets non-uniformly across layers, such as PyramidKV \citep{zhang2024pyramidkv}, or select tokens by query-aware relevance \citep{ge2024model}. 

Unlike these isolated heuristics or rigid sliding-window policies \citep{xiao2024streaming}, our theoretical framework unifies KV cache eviction under a single power-law score family parameterized by the decay exponent $\alpha$. We prove that H2O ($\alpha = 0$) and pure sliding windows ($\alpha \to \infty$) are merely boundary extremes of this continuum. Crucially, because $\alpha$ is governed by architectural dilution and relay dynamics rather than corpus-specific statistics, it exhibits strong cross-dataset generalizability—allowing a single measured $\alpha$ to serve as a reusable parameter across the datasets and tasks we test.
\section{NoHairCache: Theory and Algorithm}
\label{sec:kv_eviction}

\subsection{The Score Family and Limiting Cases}
\label{sec:score-family}

\paragraph{The admissible range of $\alpha$.} To establish the valid range of $\alpha$, we first formalize how a prompt's effect decays over generation steps. 

Following \citet{xiang2026nohair}, we take two minimally differing prompts $x,x'$ and feed each with the \emph{same} continuation~$c$. At layer $l$ and generation step $t$, the \emph{prompt offset} is defined as
\[
\Delta_l(t;c)=h_l(x'\circ c,t)-h_l(x\circ c,t),
\]
which can be considered as the difference between the two residual-stream hidden states at the same position. Its norm $\|\Delta_l(t;c)\|$ measures how much of the prompt still affects the model's internal state after $t$ steps. In this sense we call it the prompt's \emph{influence}. To form an influence profile comparable across coordinates, we average the offset over continuations, normalize each coordinate by its spread across continuations in the first window, and take the root mean square:
\[
\mathcal{S}_l(t)=\left[\frac{1}{d}\sum_{j=1}^{d}\frac{\big(\mathbb{E}_c\,\Delta_{l,j}(t;c)\big)^2}{\sigma_{l,j}^2}\right]^{1/2},
\]
where $\Delta_{l,j}$ is coordinate $j$ of $\Delta_l$, $d$ is the hidden width, $\mathbb{E}_c$ averages over the continuations shared by the prompt pair, and $\sigma_{l,j}$ is the standard deviation of $h_{l,j}$ across those continuations in the first window (pooled over both prompts, so no coordinate dominates by scale). This profile decays approximately as a power law, that is, $\mathcal{S}_l(t)\propto t^{-\alpha}$. We give now the no-hair result of \citet{xiang2026nohair}:

\begin{theorem}[No-hair bound, \citet{xiang2026nohair}]
\label{thm:nohair}
For trained Transformer language models, the power-law decay exponent of token influence satisfies $\alpha \in [0.2,\,0.7]$ for the bulk of layers asymptotically in depth while boundary layers of finite-depth models may lie outside this interval.
\end{theorem}

The bound is therefore a statement about the bulk of layers rather than every layer: measured bulk ranges agree with it, while boundary layers in some models exceed it (Table~\ref{tab:alpha-scale}).

\begin{remark}Here, the word ``trained'' means the released checkpoints our calibration script loads (Pythia-410M/1B at \texttt{step143000}, the other three at default revisions). Models with randomly initialized weights do not show this property. However, w emake no claim about partially trained model. Refer to~ \citet{xiang2026nohair} for more details.  \end{remark}

Note that Theorem~\ref{thm:nohair} constrains $\alpha$ \emph{within} the power-law family but does not select the family itself (Sec.~\ref{sec:power-law}). Its role in this paper is thus motivational: it tells us which range of $\alpha$ is worth measuring. Whether and where a given model realizes an exponent in this range remains an empirical question and we measure it across five models in Sec.~\ref{sec:alpha-scale} and confirm the measured value against downstream PPL in Sec.~\ref{sec:alpha-sweep}.

To bridge our theoretical dilution results with practical KV cache management, we introduce a unified view of eviction policies.
Specifically, for the $h$-th attention head in layer $l$, the attention matrix is
\begin{equation}
A^{(l,h)}=\mathrm{softmax}\!\left(\frac{Q^{(l,h)}K^{(l,h)\top}}{\sqrt{d_k}}+M\right),
\end{equation}
where $d_k$ is the head dimension and $M$ the causal mask. We write $a^{(l,h)}_\tau(i):=A^{(l,h)}_{\tau,i}$ for the weight assigned by the query at position $\tau$ to the key at position $i$. The cumulative historical importance of token $i$ at step $t$ is $A^{(l,h)}_i(t)=\sum_{\tau=i}^{t}a^{(l,h)}_\tau(i)$, and its age is $d_i(t)=t-i$, the number of decoding steps elapsed since token $i$ was generated.

\begin{figure}[ht]
\centering
\includegraphics[width=\linewidth]{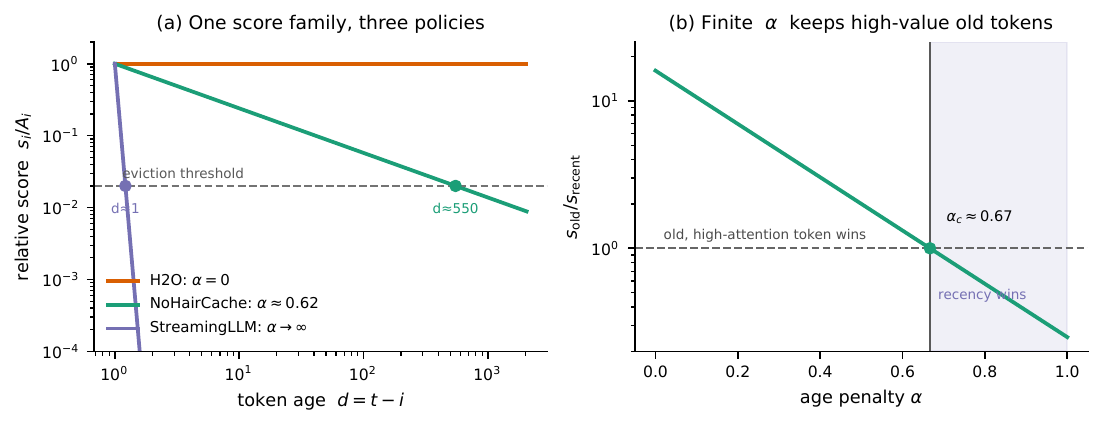}
\caption{One score family, three policies. (a) The relative score $s_i/A_i=(t-i)^{-\alpha}$ versus token age, from the H2O endpoint ($\alpha=0$) through a finite power law to the StreamingLLM endpoint ($\alpha\to\infty$). Markers show where a fixed eviction threshold is crossed. (b) A finite $\alpha$ lets an old, high-attention token outrank a recent, low-attention one until the crossover $\alpha_c=\ln(A_{\rm old}/A_{\rm recent})/\ln(d_{\rm old}/d_{\rm recent})$ with $A_{\rm old}/A_{\rm recent}=16$ and $d_{\rm old}/d_{\rm recent}=64$, $\alpha_c\approx0.67$, inside the measured range.}
\label{fig:family}
\end{figure}

We now propose the unified score family for KV-cache eviction that combines historical importance with distance-induced dilution. In its most general form the exponent may depend on the layer,
\begin{equation}
s^{(l,h)}_i(t) = A^{(l,h)}_i(t)\cdot(t-i)^{-\alpha^{(l)}},
\qquad \alpha^{(l)}\ge 0.
\end{equation}
A single scalar $\alpha$ is the special case $\alpha^{(l)}=\alpha$ for all layers.  Throughout the theoretical discussion $\alpha$ denotes a generic exponent and every statement below holds pointwise for each layer upon setting $\alpha=\alpha^{(l)}$. Per-layer $\alpha^{(l)}$ is the default while scalar $\alpha$ is used for analysis and as a deployment tier. Eviction is per-head, unlike StreamingLLM. Two axes are separate: age-penalty shape ($\alpha=0$ vs.\ $\alpha\to\infty$) and exponent granularity (scalar vs.\ per-layer). Here, $A^{(l,h)}_i(t)$ captures the accumulated historical importance, while $(t-i)^{-\alpha^{(l)}}$ (or $(t-i)^{-\alpha}$ in the scalar case) models the accessibility penalty under dilution. This family seamlessly connects existing heuristics (Figure~\ref{fig:family}) and the two propositions below are stated for a uniform exponent and hold for each layer pointwise upon setting $\alpha=\alpha^{(l)}$:

\begin{proposition}[$\alpha=0$ recovers the H2O scoring term]
For any sequence length $t$ and token position $i < t$, setting $\alpha = 0$ yields $s_i(t;0) = A_i(t)$, the Heavy-Hitter Oracle (H2O) scoring term.
\end{proposition}
\begin{proof}
The proof comes directly from the definition: $s_i(t;0) = A_i(t) \cdot 1 = A_i(t)$.
\end{proof}

\begin{proposition}[$\alpha \to \infty \implies$ Sliding Window]
\label{prop:sliding_window}
Assume the historical importance ratio $A_i(t)/A_j(t)$ is bounded in $(0,\infty)$ for all $i \neq j$. There exists a finite threshold $\alpha^* > 0$ such that for all $\alpha > \alpha^*$, the score $s_i(t;\alpha)$ is strictly monotonic with respect to token age, recovering a pure sliding window policy.
\end{proposition}
\begin{proof}
For any two positions $i < j$, we have $t-j < t-i$. The score ratio can be written as:
\begin{equation}
\frac{s_i(t;\alpha)}{s_j(t;\alpha)} = \frac{A_i(t)}{A_j(t)} \cdot \rho_{ij}^{\alpha},
\end{equation}
where $\rho_{ij} = \frac{t-j}{t-i} \in (0, 1)$. Taking the natural logarithm of both sides:
\begin{equation}
\ln \left( \frac{s_i(t;\alpha)}{s_j(t;\alpha)} \right) = \ln \left( \frac{A_i(t)}{A_j(t)} \right) + \alpha \ln \rho_{ij}.
\end{equation}
Since $\rho_{ij} < 1$, we have $\ln \rho_{ij} < 0$. As $\alpha \to \infty$, the second term dominates, driving the logarithm to $-\infty$, and thus $\frac{s_i}{s_j} \to 0$. Consequently, for any fixed pair $i < j$, there exists a threshold $\alpha_{ij} = \frac{\ln(A_i/A_j)}{-\ln \rho_{ij}}$ beyond which $s_i < s_j$. Setting $\alpha^* = \max_{i < j} \alpha_{ij}$ ensures that older tokens are strictly penalized for all $\alpha > \alpha^*$.
\end{proof}

\begin{remark}[The Attention Sink is Orthogonal]
According to Proposition~\ref{prop:sliding_window}, taking the limit $\alpha \to \infty$ gives a pure sliding window without an attention sink, because a power-law score inherently penalizes position $0$ as $t^{-\alpha} \to 0$. Thus the scoring component of StreamingLLM corresponds to the $\alpha \to \infty$ endpoint, but the full policy additionally hard-codes an attention-sink patch; a single value of $\alpha$ alone cannot reproduce that mechanism. We therefore treat $\alpha$ as a continuous parameter for decay shape and hard-coded sinks as orthogonal architectural patches where $\alpha$ is uniform across layers.
\end{remark}

\subsection{Why Power-Law over Exponential Decay} \label{sec:power-law}
Notice that we parameterize token-value decay within the power-law family $s(t)=t^{-\alpha}$ rather than the exponential family $e^{-\lambda t}$. This is a modeling choice rather than a consequence of the no-hair theorem: the theorem constrains the exponent $\alpha\in[0.2,0.7]$ \emph{within} the power-law family. We choose the power-law family for two reasons.
\begin{itemize}
    \item \textbf{No characteristic length.} Exponential decay introduces a characteristic scale $1/\lambda$ beyond which token value collapses. This would impose an artificial ``age cutoff'' on attention dilution. Attention dilution in long contexts exhibits no such natural cutoff. Instead, a token's relevance decays gradually without a preferred time scale. Power-law decay is scale-free and matches this natural behavior. For example, setting $1/\lambda=100$ causes exponential weights to plummet by nearly four orders of magnitude from $t=100$ to $t=1000$ ($e^{-1}\!\approx\!0.37$ vs.\ $e^{-10}\!\approx\!4.5\times 10^{-5}$), whereas a power law with $\alpha=0.5$ decays by only a factor of $\sqrt{10}\!\approx\!3.16$ across the same tenfold span.
    \item \textbf{Tail-safe retention.} Because exponential decay collapses rapidly, old tokens are effectively zeroed out and may be evicted even when they still carry residual global context. Power-law decay falls slowly, so an old-but-relevant token survives eviction as long as its score remains above the budget threshold. For instance, early anchor tokens, such as system prompts, entity definitions, or task constraints defined thousands of steps ago, retain non-negligible scores under a heavy tail rather than being purged in favor of transient local context.
\end{itemize}

\subsection{Calibration and Eviction Algorithm}
\label{sec:algorithm}

The decay exponents are an intrinsic property of the model, not of the dataset or task, so we can calibrate it once and reuse for other dataset or tasks. On this basis, we formulate \textbf{NoHairCache} as a two-stage procedure of offline calibration and online KV-cache eviction (Algorithm~\ref{alg:nohaircache}, Figure~\ref{fig:pipeline}). Following \citet{xiang2026nohair}, for each pair of minimally differing prompts with a shared continuation $c$ we record the prompt offset $\delta_l(t;c)=\|h_l(x'\circ c,t)-h_l(x\circ c,t)\|$, aggregate the squared offsets over prompts and continuations into a root-mean-square influence profile $\mathcal{S}_l(t)$ over logarithmically spaced (``octave'') windows $t\in\{16,32,\dots,1024\}$, and fit $\alpha^{(l)}=-\mathrm{slope}\{\log\mathcal{S}_l(t),\log t\}$ (since the influence decays as $\mathcal{S}_l(t)\propto t^{-\alpha^{(l)}}$, the log--log slope is $-\alpha^{(l)}$). 

\begin{algorithm}[t]
\caption{NoHairCache: Offline Calibration and Online KV Eviction}
\label{alg:nohaircache}
\begin{algorithmic}[1]
\Require Model with $L$ layers and $H$ heads per layer, calibration prompt pairs $\mathcal{P}_{\text{cal}}$ (default $100$ pairs per axis across $4$ axes, each pair being two minimally differing prompts with $8$ shared continuations), KV budget $B$, optional preset $\alpha_{\text{fixed}}$.
\Ensure Frozen per-layer exponents $\{\alpha^{(l)}\}_{l=1}^L$ (or a scalar $\alpha$) and evicted KV cache at each decoding step $t$.

\Statex \textbf{Stage 1: Offline Calibration (Run Once per Model)}
\If{$\alpha_{\text{fixed}}$ is not specified}
    \State Run forward passes on $\mathcal{P}_{\text{cal}}$ with shared continuations to record prompt offsets $\delta_l(t;c)$ and the influence profile $\mathcal{S}_l(t)$ across octave windows.
    \State Fit $\alpha^{(l)}=-\mathrm{slope}\{\log\mathcal{S}_l(t),\log t\}$ (or aggregate to a scalar $\alpha$).
\Else
    \State Set $\alpha \leftarrow \alpha_{\text{fixed}}$ (e.g., $\alpha = 1/2$).
\EndIf
\State Freeze the exponents in the model configuration; precompute per-layer age-decay tables $D_l[d] = d^{-\alpha^{(l)}}$ for $d \ge 1$ (a shared table $D[d]=d^{-\alpha}$ in the scalar case).

\Statex \textbf{Stage 2: Online Eviction (At Decoding Step $t$)}
\For{layer $l = 1,\dots,L$ and head $h = 1,\dots,H$}
    \State Update cumulative attention score $A^{(l,h)}_i(t)$ for active cache positions $i \le t$.
    \If{cache size exceeds budget $B$}
        \For{each candidate token $i$ in cache}
            \State Compute score $s^{(l,h)}_i(t) \leftarrow A^{(l,h)}_i(t) \cdot D_l[t - i]$.
        \EndFor
        \State Evict lowest-scoring tokens (subject to any fixed sink/window policy if configured) to maintain budget $B$.
    \EndIf
\EndFor
\end{algorithmic}
\end{algorithm}

\usetikzlibrary{arrows.meta, positioning, calc, shadows.blur}
\begin{figure}[ht]
\centering
\resizebox{\linewidth}{!}{
\begin{tikzpicture}[
    >={Stealth[length=2.2mm, width=1.6mm]},
    font=\scriptsize\sffamily, 
    node distance=5mm and 5.5mm,
    box1/.style={
        rectangle, rounded corners=3.5pt,
        draw=blue!55!black, line width=0.6pt,
        top color=blue!3, bottom color=blue!10,
        text width=34mm, minimum height=15mm,
        align=center, inner sep=3pt,
        drop shadow={shadow xshift=0.4mm, shadow yshift=-0.4mm, opacity=0.12}
    },
    box2/.style={
        rectangle, rounded corners=3.5pt,
        draw=teal!65!black, line width=0.6pt,
        top color=teal!3, bottom color=teal!10,
        text width=34mm, minimum height=15mm,
        align=center, inner sep=3pt,
        drop shadow={shadow xshift=0.4mm, shadow yshift=-0.4mm, opacity=0.12}
    },
    lbl/.style={font=\footnotesize\bfseries, anchor=west},
    ar1/.style={->, line width=0.8pt, draw=blue!55!black},
    ar2/.style={->, line width=0.8pt, draw=teal!65!black},
]

\node[box1] (cal) {Calibration pairs $\mathcal{P}_{\rm cal}$\\(100 prompt pairs)};
\node[box1, right=of cal] (fwd) {Forward passes; record\\$\delta_l(t;c)=\|h_l(x'\!\circ c,t)-h_l(x\!\circ c,t)\|$};
\node[box1, right=of fwd] (fit) {Aggregate $\mathcal{S}_l(t)$ over octaves\\$16,\dots,1024$; fit $\alpha^{(l)}=-\mathrm{slope}\{\log\mathcal{S}_l,\log t\}$};
\node[box1, right=of fit] (frz) {Freeze $\{\alpha^{(l)}\}$;\\precompute $D_l[d]=d^{-\alpha^{(l)}}$};

\node[lbl, text=blue!55!black, above=3mm of cal.north west] (s1) {Stage 1 --- Offline Calibration (once per model)};

\draw[ar1] (cal) -- (fwd);
\draw[ar1] (fwd) -- (fit);
\draw[ar1] (fit) -- (frz);

\node[box2, below=17mm of cal] (upd) {Update cumulative\\attention $A^{(l,h)}_i(t)$};
\node[box2, right=of upd] (scr) {Score candidates\\$s^{(l,h)}_i(t)=A^{(l,h)}_i(t)\,D_l[t-i]$};
\node[box2, right=of scr] (evc) {Evict lowest $s_i$\\(keep sinks / window)};
\node[box2, right=of evc] (out) {Compressed KV cache\\for the next step};

\node[lbl, text=teal!65!black, above=3mm of upd.north west] (s2) {Stage 2 --- Online Eviction (at every decoding step $t$)};

\draw[ar2] (upd) -- (scr);
\draw[ar2] (scr) -- (evc);
\draw[ar2] (evc) -- (out);

\draw[ar2, dashed, rounded corners=4pt] 
    (frz.south) -- ++(0,-4.5mm) -| 
    node[pos=0.25, fill=white, inner sep=1.5pt, font=\tiny\itshape, text=teal!65!black] {reuse frozen $D_l[d]$} 
    (scr.north);

\node[
    rectangle, rounded corners=2pt, fill=gray!8, draw=gray!30,
    inner xsep=8pt, inner ysep=3pt, font=\scriptsize,
    below=5mm of $(scr.south)!0.5!(evc.south)$
] (spectrum) {
    \textbf{Unified Spectrum (uniform $\alpha$):} \quad $\alpha = 0$ (H2O) \quad $\longleftarrow\!\!-\!\!\!-\!\!\longrightarrow$ \quad $\alpha \to \infty$ (StreamingLLM)
};

\end{tikzpicture}%
}
\caption{NoHairCache pipeline corresponding to Algorithm~\ref{alg:nohaircache}. Stage 1 runs once per model: prompt offsets $\delta_l(t;c)$ are aggregated into influence curves $\mathcal{S}_l(t)$, whose log--log slope gives the per-layer exponent via $\alpha^{(l)}=-\mathrm{slope}(\log\mathcal{S}_l,\log t)$. The frozen exponents and per-layer age tables $D_l[d]=d^{-\alpha^{(l)}}$ are reused at every decoding step. Stage 2 scores cached tokens with $s^{(l,h)}_i(t)=A^{(l,h)}_i(t)\,D_l[t-i]$ and evicts the lowest-scoring ones.}
\label{fig:pipeline}
\end{figure}
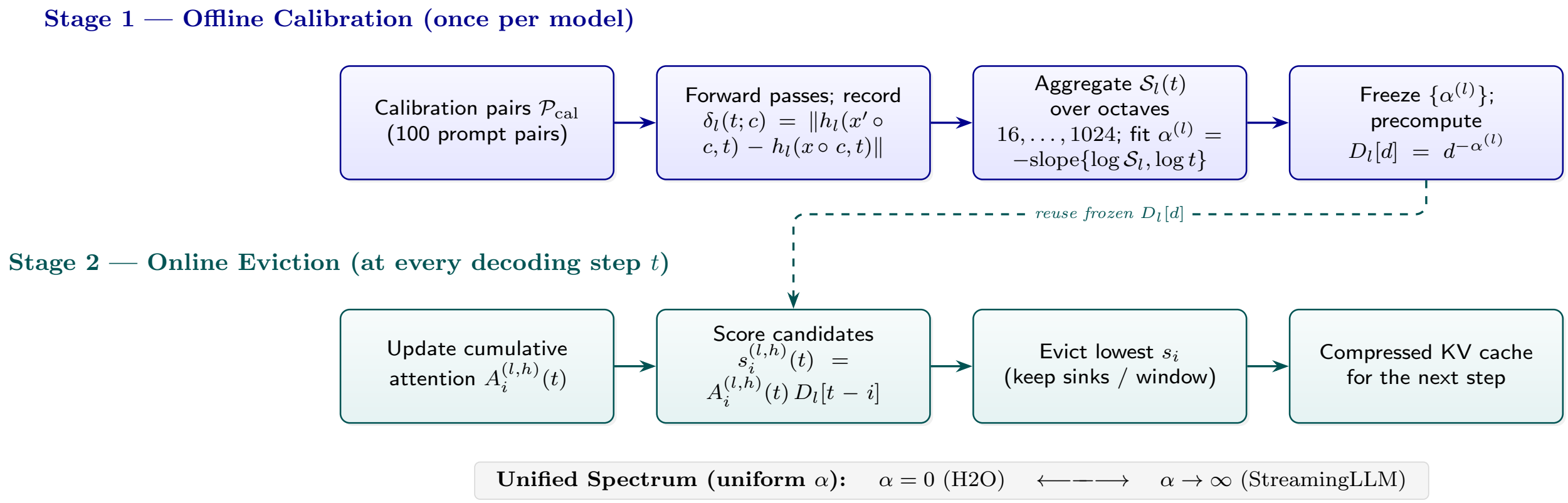

\paragraph{Online efficiency and deployment tiers.}
Once the exponents are frozen, precomputing $d^{-\alpha^{(l)}}$ into per-layer age-indexed lookup tables avoids repeated exponentiation during decoding: scoring a token requires only one table lookup and one multiplication. Compared to H2O ($\alpha = 0$), the incremental online overhead is just one multiplication per scored entry at eviction steps. In practice, the framework supports three deployment tiers under a single configuration switch: (i) the theoretical default $\alpha = 1/2$ with zero calibration cost, (ii) a single frozen scalar $\alpha$ calibrated offline, and (iii) per-layer $\alpha^{(l)}$ calibrated offline for finer layer-wise granularity.

\paragraph{Calibration is not hyperparameter tuning.}
Like LayerNorm's running statistics, $\alpha$ is a structural model quantity estimated once and read at inference, not tuned per downstream task. We calibrate once per model on WikiText-103 and keep it frozen across all datasets, budgets, and benchmarks.
\section{Experiments and Results}
\label{sec:experiments}

In this section, we evaluate the \textsc{NoHairCache} algorithm introduced in Sec.~\ref{sec:algorithm}. We first describe the setup, then examine the measured exponents, verify them against perplexity sweeps, and test cross-dataset transfer, budget regimes, cross-model robustness, and downstream QA.

\subsection{Setup}
\label{sec:setup}

\paragraph{Models and calibration.}
We evaluate on autoregressive language models from the Pythia and Qwen families, including Pythia-410M, Pythia-1B, Pythia-1.4B, Qwen2.5-0.5B, and Qwen2.5-1.5B-Instruct (denoted Qwen2.5-1.5B below). All $\alpha$ calibrations are performed once on the WikiText-103 validation split using $100$ prompt pairs per axis across $4$ axes (detail, para, lang, unrelated). Each pair consists of two minimally differing prompts, and both prompts are teacher-forced with the same $8$ shared continuations, i.e.\ $16$ forward passes per pair and $4\times100\times16=6{,}400$ passes in total (the subsequent bootstrap resampling and power-law fitting take only a few seconds). On a MacBook with an Apple M3 chip and $36\,\text{GB}$ unified memory (MPS backend, octave windows up to $1024$ tokens), a full calibration run takes approximately $20.3\,\text{min}$ for Pythia-410M ($23$ layers), $49.8\,\text{min}$ for Pythia-1B ($15$ layers), $67.4\,\text{min}$ for Pythia-1.4B ($24$ layers), and $78.6\,\text{min}$ for Qwen2.5-1.5B ($27$ layers), scaling roughly linearly with the number of pairs and continuations and
with maximum window length (timings reported for four of the five models and
Qwen2.5-0.5B calibration follows the same procedure). However, our framework scales  to larger models and context windows for readers with greater compute budgets.

\paragraph{Evaluation protocol.}
We evaluate perplexity across WikiText-103, PG19, and arXiv on $50$ sequences of $2{,}048$ tokens per dataset with a streaming chunk size of $128$. KV-cache eviction is simulated via an attention mask, which is mathematically equivalent to physically cropping key and value entries while avoiding RoPE position-index bookkeeping artifacts.

\subsection{Measured \texorpdfstring{$\alpha$}{alpha} across Models and Scales}
\label{sec:alpha-scale}

Table~\ref{tab:alpha-scale} reports the measured exponent across five
models and two architectures: three Pythia variants (410M, 1B, 1.4B) and two
Qwen variants (0.5B, 1.5B), spanning a $3\times$ parameter range. 

The
per-model mean $\alpha$ lies between $0.602$ and $0.663$ (coefficient of
variation under $4\%$), and the per-model bulk intervals all fall inside
$[0.533,0.676]$. Two observations follow. First, $\alpha$ does not drift
with scale: the three Pythia means are $0.637$, $0.663$, and $0.634$ for
410M, 1B, and 1.4B, and the two Qwen means are $0.602$ and $0.620$ for 0.5B
and 1.5B---the Pythia means are non-monotonic in scale, and the Qwen mean
rises by only $0.018$. Second, $\alpha$ does not drift with
architecture: the Qwen means are slightly lower than the Pythia means
($0.602$--$0.620$ vs.\ $0.634$--$0.663$) but still fall within the Pythia
per-layer range.

\begin{figure}[ht]
\centering
\includegraphics[width=0.9\linewidth]{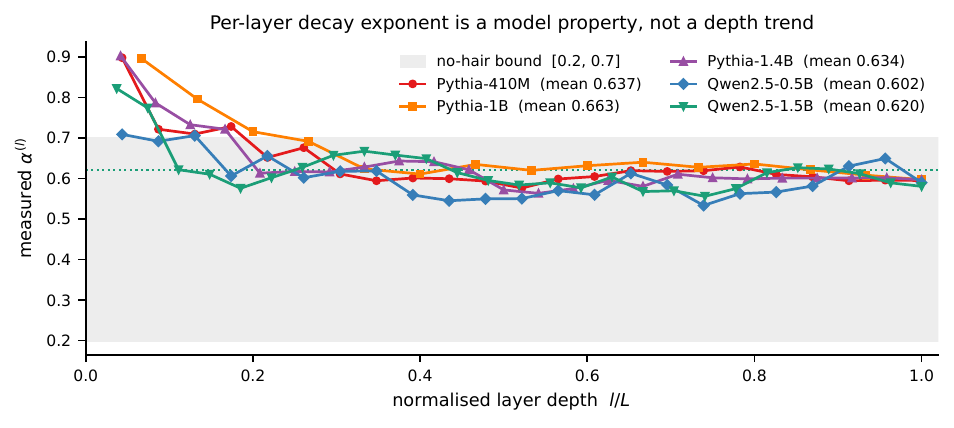}
\caption{Measured per-layer exponent $\alpha^{(l)}$ (hidden-state influence, detail axis) for all five models, plotted against normalised depth $l/L$. The means span only $0.602$--$0.663$, and the no-hair interval $[0.2,0.7]$ is shaded. Boundary layers carry the highest exponent in every model while the middle layers are comparatively flat, as discussed in Sec.~\ref{sec:alpha-scale}.}
\label{fig:layers}
\end{figure}

 The layer
structure is only partly preserved (Figure~\ref{fig:layers}): in every model the first two layers carry
the highest exponent and the middle layers are comparatively flat, but the
final layers do not show a consistent rebound across models; the full
per-layer profile for Qwen2.5-1.5B is listed in Table~\ref{tab:alpha-layers}
(Appendix~\ref{app:alpha}).

The one systematic scale effect we observe is in the \emph{upper} end of the
range: Qwen-0.5B, the smallest model, has an upper bound of only $0.709$,
barely above the no-hair bound of $0.7$, whereas Qwen-1.5B reaches $0.822$.
For Pythia, by contrast, the upper bound is flat across scales
($0.896$--$0.902$). We have no explanation for this architecture-dependent
difference; note that Qwen-0.5B and Pythia-410M have the same depth (23
measured layers) yet very different upper bounds. The mean
$\alpha$, a scalar summary of the calibrated profile, is unaffected by this effect and
stays within $0.602$--$0.663$ across the full scale range.

We read these observations as evidence that the mean $\alpha$ is a property
of the training dynamics rather than of a particular depth, width, or
architecture: depth varies from $15$ to $27$ across our models, width from
$896$ to $2048$, yet the mean exponent moves by less than $0.07$ in absolute terms. This
is consistent with the attention-dilution mechanism of
\citet{xiang2026nohair}, which is a property of the attention operation
itself and should not depend on the specific architecture.

\begin{table}[ht]
    \centering
    \caption{Measured decay exponent across five models and two architectures. ``Bulk'' is the range of the middle layers after removing the first and last four measured layers; layers without a valid fit are excluded, so the layer count may differ from the official depth by one.}
    \label{tab:alpha-scale}
    \begin{tabular}{llS[table-format=1.3]cc}
        \toprule
        {Architecture} & {Model} & {$\alpha$ mean} & $\alpha$ range & Bulk \\
        \midrule
        Pythia & 410M & 0.637 & $[0.576, 0.898]$ & $[0.576, 0.676]$ \\
        Pythia & 1B   & 0.663 & $[0.597, 0.896]$ & $[0.611, 0.640]$ \\
        Pythia & 1.4B & 0.634 & $[0.563, 0.902]$ & $[0.563, 0.643]$ \\
        \midrule
        Qwen   & 0.5B & 0.602 & $[0.533, 0.709]$ & $[0.533, 0.656]$ \\
        Qwen   & 1.5B & 0.620 & $[0.556, 0.822]$ & $[0.556, 0.667]$ \\
        \bottomrule
    \end{tabular}
\end{table}

\subsection{The Measured \texorpdfstring{$\alpha$}{alpha} Agrees with the PPL Optimum}
\label{sec:alpha-sweep}

\begin{figure}[ht]
\centering
\includegraphics[width=0.62\linewidth]{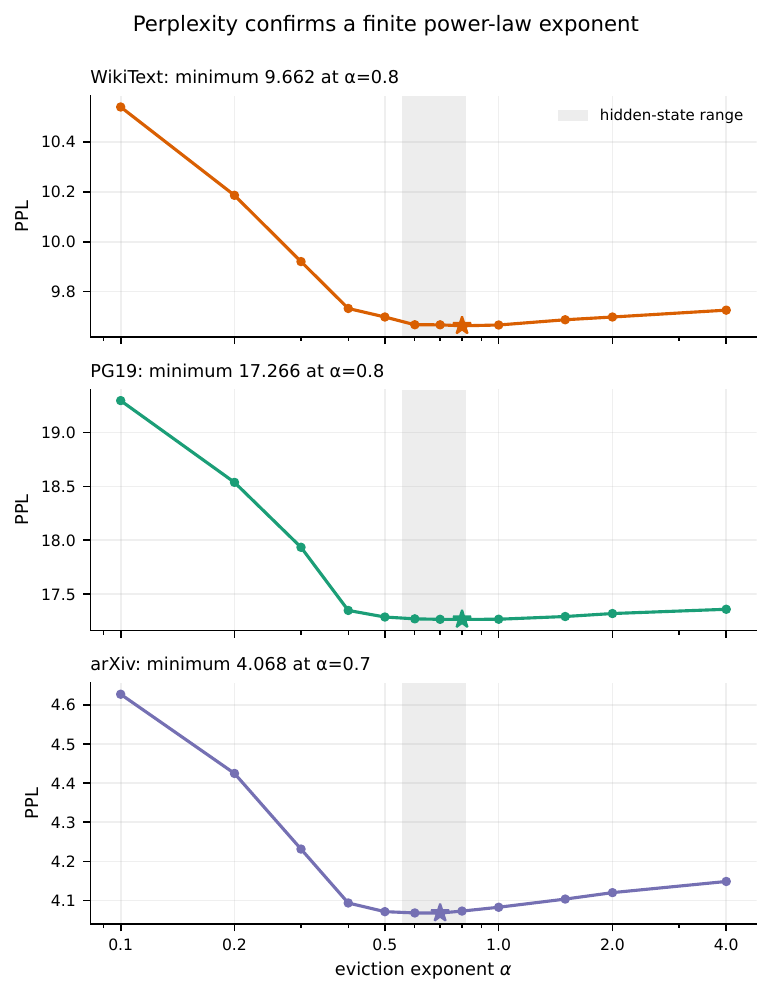}
\caption{$\alpha$-sweep perplexity at budget $512$ on all three datasets (same calibration, diagnostic only). The grey band marks the hidden-state measurement range $[0.556,0.822]$ and stars mark each dataset's minimum. PPL is flat for $\alpha\in[0.6,0.8]$ and rises for $\alpha>1$, so the optimum is a finite power-law exponent rather than a boundary artifact. This is the raw data behind Table~\ref{tab:scan}.}
\label{fig:sweep}
\end{figure}

Before evaluating the full system, we verify that the measured $\alpha$ is
the right value to deploy. An $\alpha$-sweep on downstream
perplexity, that is, a diagnostic sweep, not a deployment-time calibration, varies
$\alpha$ from $0.1$ to $4.0$ on all three evaluation datasets
(Figure~\ref{fig:sweep}, Table~\ref{tab:scan}). The PPL-optimal exponent falls in $[0.6,0.8]$ on
every dataset, overlapping the hidden-state measurement $[0.556,0.822]$ of
Sec.~\ref{sec:alpha-scale}. The curve is flat on this interval and
rises monotonically for $\alpha>1$, so the plateau is a genuine optimum
rather than a boundary artifact of the sweep range. The coarse $0.1$-step
grid resolves the arXiv minimum to $4.068$, and refining $\alpha$ or using
the per-layer vector lowers it by a further $0.002$ PPL
(Table~\ref{tab:calib-ablation}) without moving the optimum. As $\alpha$ grows, PG19
PPL approaches the StreamingLLM endpoint ($17.360$ at $\alpha=4.0$ versus
$17.395$), confirming the $\alpha\to\infty$ limit of
Proposition~\ref{prop:sliding_window} empirically while remaining far below
the H2O endpoint. Three datasets and the hidden-state measurement therefore
agree on where $\alpha$ lives, and all are consistent with a finite
power-law exponent without assuming a specific range. The sweep uses a
global scalar for clarity and the per-layer variant used in
Table~\ref{tab:main} differs by at most $0.04$ PPL (Table~\ref{tab:calib-ablation}). The full NoHairCache pipeline, with
$\alpha$ frozen at the measured value, is evaluated below.

\begin{table}[ht]
\centering
\caption{$\alpha$-sweep on WikiText validation, 50 sequences, budget $=512$,
evaluated on all three datasets with the same calibration. PPL is flat on
$[0.6,0.8]$ and rises monotonically for $\alpha>1$ on every dataset. The
three per-dataset minima (WikiText and PG19 at $\alpha=0.8$, arXiv at
$\alpha=0.6$--$0.7$ on this grid, to within the $0.002$ PPL refinement in
Table~\ref{tab:calib-ablation}) all lie inside the hidden-state measurement
range $[0.556,0.822]$. This is a diagnostic sweep, not a deployment-time
calibration: it varies a single frozen $\alpha$ across all layers.}
\label{tab:scan}
\begin{tabular}{S[table-format=1.1] S[table-format=2.3] S[table-format=2.3] S[table-format=1.3]}
\toprule
{$\alpha$} & {WikiText} & {PG19} & {arXiv} \\
\midrule
0.1 & 10.541 & 19.297 & 4.628 \\
0.2 & 10.186 & 18.538 & 4.425 \\
0.3 & 9.920  & 17.935 & 4.231 \\
0.4 & 9.732  & 17.349 & 4.093 \\
0.5 & 9.698  & 17.288 & 4.071 \\
0.6 & 9.666  & 17.271 & \bfseries 4.068 \\
0.7 & 9.666  & 17.267 & \bfseries 4.068 \\
0.8 & \bfseries 9.662 & \bfseries 17.266 & 4.073 \\
1.0 & 9.666  & 17.268 & 4.083 \\
1.5 & 9.687  & 17.293 & 4.103 \\
2.0 & 9.698  & 17.321 & 4.120 \\
4.0 & 9.725  & 17.360 & 4.149 \\
\bottomrule
\end{tabular}
\end{table}

\paragraph{How sensitive is the result to how $\alpha$ is set?}
Table~\ref{tab:calib-ablation} varies the calibration strategy while holding
everything else fixed. A single global scalar matched to the measured mean
($\alpha=0.61$), the per-layer vector, and a randomly shuffled per-layer
vector all land within $0.04$ PPL of each other on every dataset. The theory
default $\alpha=0.5$ is slightly worse than the global-scalar and shuffled
variants on all three datasets. On PG19 it is marginally better than the
per-layer row ($17.288$ vs $17.305$), which itself sits slightly above the
other calibrated variants. The result is therefore insensitive to the
granularity of calibration: a single scalar matched to the measured mean
suffices, and the per-layer structure contributes only a second-order
correction. We report the per-layer variant throughout as the principled
choice, noting that on this fixed evaluation slice the differences among
calibrated variants are small and do not change the conclusions, and that the
global scalar is in fact slightly better on WikiText and PG19. This is consistent with $\alpha$
being a coarse-grained model property rather than a per-layer tuning knob.

\begin{table}[ht]
\centering
\caption{Calibration sensitivity at budget $=512$, Qwen2.5-1.5B.
All rows use the same score family, only how $\alpha$ is obtained differs.
Global, per-layer, and shuffled per-layer are within $0.04$ PPL. The theory
default $\alpha=0.5$ is slightly worse than the global and shuffled variants
on every dataset. ``Shuffled
per-layer'' randomly permutes the $\alpha^{(l)}$ vector across layers.}\label{tab:calib-ablation}
\begin{tabular}{lS[table-format=1.3]S[table-format=2.3]S[table-format=1.3]}
\toprule
{Calibration} & {WikiText} & {PG19} & {arXiv} \\
\midrule
Theory default ($\alpha=0.5$) & 9.698 & 17.288 & 4.071 \\
Global scalar ($\alpha=0.61$) & 9.664 & 17.271 & 4.066 \\
Per-layer $\alpha^{(l)}$      & 9.691 & 17.305 & 4.066 \\
Shuffled per-layer            & 9.672 & 17.283 & 4.066 \\
\bottomrule
\end{tabular}
\end{table}

\subsection{Cross-Dataset Generalization}
\label{sec:results_generalization}

The comparison is stable across corpora: with $\alpha$ frozen from WikiText
at budget $=512$ (Table~\ref{tab:main}), NoHairCache attains the lowest
perplexity on every dataset, outperforming StreamingLLM ($\alpha=\infty$) by
$0.057$/$0.090$/$0.104$ PPL on WikiText/PG19/arXiv and also beating H2O
($\alpha=0$).
Against StreamingLLM: WikiText $9.691$ vs $9.748$ ($p=0.0059$), PG19
$17.305$ vs $17.395$ ($p=0.0218$), arXiv $4.066$ vs $4.170$ ($p<0.001$).
Against H2O the margins are larger ($-1.07$, $-2.39$, and $-0.72$ PPL
respectively). A paired Wilcoxon signed-rank test over the 50 sequences
confirms both comparisons on every dataset: NoHairCache beats StreamingLLM
on $34/33/44$ of 50 sequences, and beats H2O on $50/50/50$ sequences
($p<10^{-14}$ on all three datasets).
We attribute H2O's poor performance to cumulative attention being dominated
by early high-frequency filler tokens, so middle-context content is crowded
out of the top-$k$ regardless of dataset.

\begin{table}[ht]
\centering
\caption{Cross-dataset generalization at budget $=512$. $\alpha$ calibrated
on WikiText only (Qwen2.5-1.5B), used unchanged on all three datasets.
NoHairCache is the lowest-PPL method on every row. $\Delta$ and $p$ are
NoHairCache vs StreamingLLM (Wilcoxon signed-rank, 50 sequences).}
\label{tab:main}
\begin{tabular}{lS[table-format=2.3]S[table-format=2.3]S[table-format=2.3]S[table-format=+1.3]l}
\toprule
{Dataset} & {H2O} & {NoHairCache} & {StreamingLLM} & {$\Delta$} & {$p$} \\
\midrule
WikiText & 10.764 & \bfseries 9.691 & 9.748 & -0.057 & 0.0059 \\
PG19 & 19.696 & \bfseries 17.305 & 17.395 & -0.090 & 0.0218 \\
arXiv & 4.786 & \bfseries 4.066 & 4.170 & -0.104 & $<$0.001 \\
\bottomrule
\end{tabular}
\end{table}

\subsection{Budget Curves}
\label{sec:results_budget}

\begin{figure}[ht]
\centering
\includegraphics[width=\linewidth]{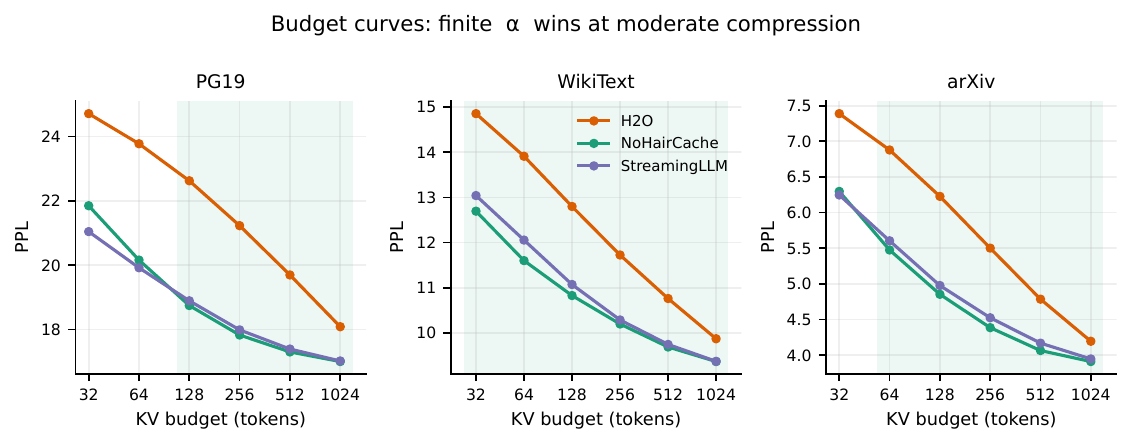}
\caption{KV-budget curves for Qwen2.5-1.5B with $\alpha$ calibrated on WikiText only (shaded: budgets where NoHairCache beats StreamingLLM). NoHairCache dominates at moderate compression, nearly ties near full cache (still slightly ahead), and loses only at the tightest budgets. The crossover shifts with the dataset (PG19 $\approx128$, arXiv $\approx64$, WikiText $\le32$). Raw numbers: Table~\ref{tab:budget} and Appendix~\ref{app:budget}.}
\label{fig:budget}
\end{figure}

Table~\ref{tab:budget} and Figure~\ref{fig:budget} report PG19 with six budgets from 32 to 1024.
NoHairCache wins when 6--25\% of tokens are retained, ties StreamingLLM near
full cache ($>50\%$ retained), and loses only below $\approx6\%$ retention
(budget $\leq64$), where the pure-recent strategy is optimal.
H2O is worst at every budget.
The crossover point moves with distribution: PG19 at $\approx128$,
WikiText at $\leq32$, and arXiv at $\approx64$
(Appendix~\ref{app:budget}).

\begin{table}[ht]
\centering
\caption{PG19 budget curve (Qwen2.5-1.5B, $\alpha$ from WikiText
calibration). The crossover between StreamingLLM and NoHairCache occurs between
64 and 128 ($\approx3$--$6\%$ retained). Below it, StreamingLLM's pure-recent
strategy wins. Above it, NoHairCache wins. Near full cache the two nearly
tie, with NoHairCache still slightly ahead. H2O is worst at every budget.}
\label{tab:budget}
\begin{tabular}{S[table-format=4.0] S[table-format=2.1] S[table-format=2.3] S[table-format=2.3] S[table-format=2.3]}
\toprule
{Budget} & {Retained (\%)} & {H2O} & {NoHairCache} & {StreamingLLM} \\
\midrule
32 & 1.5 & 24.705 & 21.845 & \bfseries 21.042 \\
64 & 3.1 & 23.770 & 20.158 & \bfseries 19.919 \\
128 & 6.2 & 22.621 & \bfseries 18.746 & 18.893 \\
256 & 12.5 & 21.227 & \bfseries 17.830 & 17.990 \\
512 & 25.0 & 19.696 & \bfseries 17.305 & 17.395 \\
1024 & 50.0 & 18.088 & \bfseries 17.006 & 17.025 \\
\bottomrule
\end{tabular}
\end{table}

\subsection{Cross-Model Verification}
\label{sec:results_crossmodel}

Table~\ref{tab:pythia} reports Pythia-1.4B on all three datasets at budgets
128 and 512, using Pythia's own per-layer $\alpha$ measured on WikiText only.
NoHairCache wins every cell on this model.
For WikiText and PG19 the margin is larger at budget 128 ($-0.23$ and
$-0.32$) than at 512 ($-0.06$ and $-0.12$), consistent with the
budget-curve finding that the advantage is largest under moderate
compression.
For arXiv the two budgets are comparable ($-0.081$ at 128, $-0.094$ at
512), consistent with arXiv's earlier crossover ($\approx64$).

\begin{table}[ht]
\centering
\caption{Cross-model verification on Pythia-1.4B. $\alpha$ calibrated on
WikiText only. NoHairCache wins every cell and the margin is larger at budget
128 than at 512, consistent with the budget curve on Qwen.}
\label{tab:pythia}
\begin{tabular}{llS[table-format=2.3]S[table-format=2.3]S[table-format=2.3]S[table-format=+1.3]}
\toprule
{Dataset} & {Budget} & {H2O} & {NoHairCache} & {StreamingLLM} & {$\Delta$} \\
\midrule
WikiText & 128 & 17.964 & \bfseries 14.180 & 14.408 & -0.228 \\
WikiText & 512 & 14.251 & \bfseries 12.367 & 12.431 & -0.064 \\
\midrule
PG19 & 128 & 19.615 & \bfseries 15.630 & 15.951 & -0.321 \\
PG19 & 512 & 16.924 & \bfseries 14.735 & 14.857 & -0.122 \\
\midrule
arXiv & 128 & 7.129 & \bfseries 5.380 & 5.461 & -0.081 \\
arXiv & 512 & 5.385 & \bfseries 4.470 & 4.564 & -0.094 \\
\bottomrule
\end{tabular}
\end{table}

To test whether the advantage depends on model capacity, we repeat the
comparison on Pythia-1B, Pythia-410M, and Qwen-0.5B (Table~\ref{tab:scale}).
The advantage over StreamingLLM grows with scale, and the pattern is
consistent across the two architectures. At 1.4B and 1.5B, NoHairCache wins
every cell ($\Delta\in[-0.38,-0.05]$). At 1B, it ties on WikiText at
budget 128 and is better or equal elsewhere ($\Delta\in[-0.11,0.00]$). At
410M and 0.5B, the picture separates by dataset: Pythia-410M is worse on
WikiText ($+0.21$ and $+0.14$) but comparable on PG19 ($-0.01$ and $+0.01$)
and better on arXiv ($-0.01$ and $-0.08$) while Qwen-0.5B is better on
WikiText-128 ($-0.31$) and on arXiv, but worse on PG19 ($+0.26$ and
$+0.04$). Against H2O, every model still improves substantially at every
budget.

The measured $\alpha$ itself is stable across these scales
(Table~\ref{tab:alpha-scale}), so the trend reflects the eviction policy's
usable signal rather than the exponent's estimation. Small models attend
more locally, so a pure sliding window is already close to optimal and the
benefit of retaining distant tokens is small. The advantage over
StreamingLLM emerges once the model has enough capacity to exploit the
retained context. The asymmetry between datasets at small scale, which is, worse on
PG19, better on WikiText, is consistent with this account: PG19's
narrative structure is the most locally predictable of the three, so a
pure-recent window is hardest to beat there. NoHairCache therefore targets
models of roughly 1B parameters and above, which is also the regime where
KV-cache eviction matters in practice. At smaller scales it remains a strict
improvement over attention-accumulation baselines but not over a well-tuned
sliding window.

\begin{table}[ht]
\centering
\caption{Cross-scale robustness: $\Delta$ = NoHairCache $-$ StreamingLLM
in aggregate PPL (negative is better). The advantage grows with scale and
the pattern is consistent across architectures: the 1.4B/1.5B models win
every cell, the three $\leq$1B models are mixed. Evaluation uses a fixed
slice of 50 non-overlapping 2048-token segments per dataset, so the
differences are deterministic rather than sampling noise. $\alpha$ is
calibrated on WikiText and frozen for every cell.}
\label{tab:scale}
\begin{tabular}{llS[table-format=+1.3]S[table-format=+1.3]S[table-format=+1.3]S[table-format=+1.3]S[table-format=+1.3]S[table-format=+1.3]}
\toprule
{Model} & {Params} & {WT-128} & {WT-512} & {PG-128} & {PG-512} & {AX-128} & {AX-512} \\
\midrule
Pythia-410M  & 410M & +0.213 & +0.140 & -0.014 & +0.012 & -0.008 & -0.083 \\
Pythia-1B    & 1.0B & +0.000 & -0.038 & -0.097 & -0.040 & -0.011 & -0.106 \\
Pythia-1.4B  & 1.4B & -0.228 & -0.064 & -0.321 & -0.122 & -0.081 & -0.094 \\
Qwen-0.5B    & 0.5B & -0.305 & -0.013 & +0.255 & +0.038 & -0.025 & -0.114 \\
Qwen-1.5B    & 1.5B & -0.226 & -0.057 & -0.379 & -0.090 & -0.104 & -0.104 \\
\bottomrule
\end{tabular}
\end{table}

\subsection{Downstream Task Accuracy}
\label{sec:results_downstream}

\begin{figure}[ht]
\centering
\includegraphics[width=\linewidth]{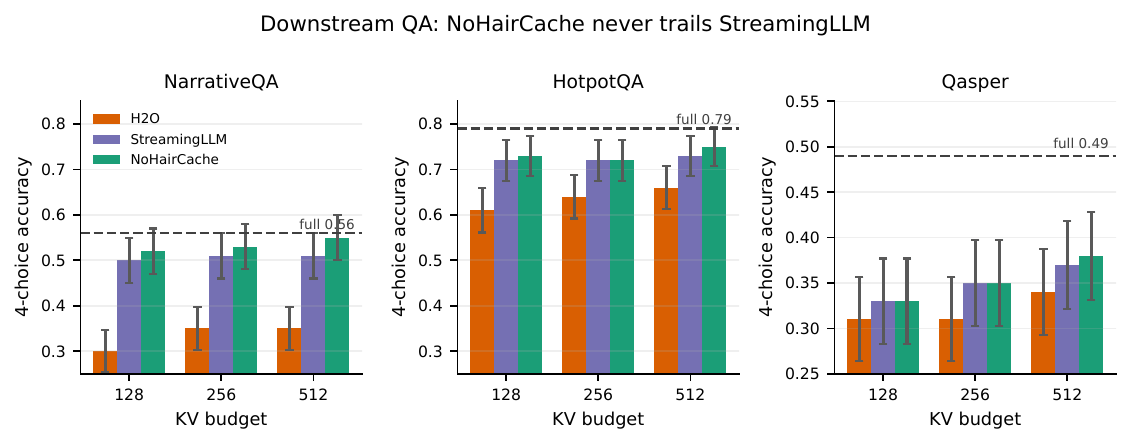}
\caption{Four-way multiple-choice accuracy on three LongBench QA subsets (random $=0.25$, 100 samples per cell, $\pm1$ s.e.), dashed lines are full-cache accuracy. NoHairCache is never worse than StreamingLLM and H2O is worst in every cell, with the largest margin on NarrativeQA at budget $512$. Raw numbers: Table~\ref{tab:longbench}.}
\label{fig:longbench}
\end{figure}

Table~\ref{tab:longbench} and Figure~\ref{fig:longbench} report four-way multiple-choice accuracy (random = 0.25) on
three LongBench QA subsets, 100 samples per cell, with $\alpha$ calibrated
on WikiText only.
NoHairCache is never worse than StreamingLLM (6 better, 3 tied, 0 worse, 
one-sided sign test $p\approx0.016$) while H2O is worst in all 9 cells.
A paired per-question McNemar test (exact binomial, two-sided) finds no
significant advantage in any individual cell: exact $p\ge0.125$ for all nine
(task, budget) cells, with Holm-adjusted $p=1.00$ throughout. The sign test above is
therefore descriptive rather than confirmatory, and we do not claim
per-task statistical significance for the accuracy gains.
The advantage is largest on NarrativeQA, the single-document task where the
answer lies in the middle of a long narrative: at budget $=512$, NoHairCache
retains $98.2\%$ of full-cache accuracy ($0.55$ vs $0.56$), whereas
StreamingLLM retains $91.1\%$ ($0.51$ vs $0.56$).
This 7-percentage-point gap is consistent with truncation of
middle-context tokens.  We do not test this mechanism directly.
On Qasper, full-cache accuracy is only $0.49$ and all methods degrade to
$0.31$--$0.38$, and the task does not discriminate between policies
(Appendix~\ref{app:longbench}).

\begin{table}[ht]
\centering
\caption{Four-way multiple-choice accuracy on LongBench QA subsets (random = $0.25$),
100 samples per cell, $\alpha$ calibrated on WikiText only.
NoHairCache is never worse than StreamingLLM (6 better, 3 tied, 0 worse,
one-sided sign test $p\approx0.016$) while H2O is worst in all 9 cells.
Per-cell exact McNemar tests are not significant after Holm correction.
The largest margin is on NarrativeQA at budget $=512$, where NoHairCache
retains $98.2\%$ of full-cache accuracy versus $91.1\%$ for StreamingLLM.}
\label{tab:longbench}
\begin{tabular}{llS[table-format=1.2]S[table-format=1.2]S[table-format=1.2]S[table-format=1.2]S[table-format=2.1]}
\toprule
{Subset} & {Budget} & {Full} & {H2O} & {StreamingLLM} & {NoHairCache} & {Retain (\%)} \\
\midrule
NarrativeQA & 128 & 0.56 & 0.30 & 0.50 & \bfseries 0.52 & 92.9 \\
NarrativeQA & 256 & 0.56 & 0.35 & 0.51 & \bfseries 0.53 & 94.6 \\
NarrativeQA & 512 & 0.56 & 0.35 & 0.51 & \bfseries 0.55 & \bfseries 98.2 \\
\midrule
HotpotQA & 128 & 0.79 & 0.61 & 0.72 & \bfseries 0.73 & 92.4 \\
HotpotQA & 256 & 0.79 & 0.64 & 0.72 & 0.72 & 91.1 \\
HotpotQA & 512 & 0.79 & 0.66 & 0.73 & \bfseries 0.75 & 94.9 \\
\midrule
Qasper & 128 & 0.49 & 0.31 & 0.33 & 0.33 & 67.3 \\
Qasper & 256 & 0.49 & 0.31 & 0.35 & 0.35 & 71.4 \\
Qasper & 512 & 0.49 & 0.34 & 0.37 & \bfseries 0.38 & 77.6 \\
\bottomrule
\end{tabular}
\end{table}

\section{Discussion and Limitations}
\label{sec:discussion}

\textbf{Why NoHairCache wins.}
The mechanism is visible in the budget curve, not in a separate retrieval
benchmark: under moderate compression, the cache has enough room to retain
some distant tokens, and a finite $\alpha$ retains exactly those whose
cumulative attention justifies their age.
Under extreme compression, there is little room for distant tokens regardless
of score, where StreamingLLM wins is dataset-specific (budgets $\le 64$ on
PG19, $32$ on arXiv, and at or below $32$ on WikiText).
Near full cache, all policies approximate full attention.
NoHairCache's advantage is therefore regime-specific and mechanistically
explained by the budget curve itself.

\textbf{Proxy limitation.}
Prior work suggests that attention-mass proxies can underestimate the decay
exponent due to softmax recency bias \citep{xiang2026nohair} and we do not
quantify the discrepancy here.
This is a property of the proxy, not of the framework: the hidden-state
measurement and the PPL sweep agree with each other.

\textbf{What we do not claim.}
We do not claim that NoHairCache preserves middle-context information better
than StreamingLLM and we have not run a retrieval benchmark that would establish
this.
We claim that (i) the decay exponent is a measurable model property,
(ii) measuring it once on one dataset generalizes to others, and
(iii) using it improves language-modeling perplexity across datasets and
models without per-dataset tuning.

\textbf{Boundary conditions.}
Our results cover models up to 1.5B parameters and 2048-token sequences.
The dilution mechanism behind the no-hair bound is scale-free, so we expect
the calibrated exponent to remain stable at larger scales, but this has not
been tested.

\section{Conclusion}
\label{sec:conclusion}

Existing KV-eviction policies implicitly choose a decay exponent: H2O sets it
to zero, StreamingLLM to infinity. We have shown that these are the two
endpoints of a single power-law score family, and that the exponent between
them is not a heuristic knob but a measurable property of the model, whose
admissible range is narrowed by the no-hair bound.

\textbf{What we established.}
The decay exponent is measurable from hidden-state influence with a one-time
calibration per model: its bulk value is stable across scale and architecture
and agrees with the optimum of an independent perplexity sweep. Once frozen on
a single corpus, the exponent transfers unchanged to unseen domains, cache
budgets, and downstream tasks, without per-dataset tuning. The resulting
drop-in scorer adds one multiplication per eviction over H2O and lowers
perplexity on WikiText, PG19, and arXiv, and improves four-way
multiple-choice accuracy on LongBench, in the regime where eviction actually
matters.

\textbf{Scope and outlook.}
The gain is regime-specific and mechanistically explained by the budget curve:
a finite exponent helps when there is enough room to retain distant tokens
whose cumulative attention justifies their age, and vanishes near full cache
or under the tightest budgets. Our results cover five models and 2048-token
sequences, and we do not claim to preserve middle-context information beyond
what perplexity and QA measure. More broadly, the no-hair view suggests that
cache compression need not be a search over heuristics: a single structural
quantity, calibrated once, may be enough to place any policy on the
accumulation--recency spectrum. Extending this view to longer contexts, other
architectures, and per-head exponents is left to future work.
\appendix
\clearpage

\section{Per-layer \texorpdfstring{$\alpha$}{alpha} (Qwen2.5-1.5B, WikiText, detail axis)}
\label{app:alpha}

\begin{table}[H]
\centering
\caption{Per-layer $\alpha^{(l)}$ measured by hidden-state influence
(teacher-forced prompt pairs, detail axis), Qwen2.5-1.5B.
The measured range $[0.556,0.822]$ is consistent with a finite power-law
exponent and matches the PPL-optimal plateau $[0.6,0.8]$ while boundary layers
(L1--L2) sit slightly higher, as expected for finite-depth models.}
\label{tab:alpha-layers}
\begin{tabular}{S[table-format=2.0] S[table-format=1.3] S[table-format=2.0] S[table-format=1.3] S[table-format=2.0] S[table-format=1.3]}
\toprule
{Layer} & {$\alpha$} & {Layer} & {$\alpha$} & {Layer} & {$\alpha$} \\
\midrule
1  & 0.822 & 10 & 0.658 & 19 & 0.570 \\
2  & 0.774 & 11 & 0.649 & 20 & 0.556 \\
3  & 0.622 & 12 & 0.615 & 21 & 0.575 \\
4  & 0.611 & 13 & 0.595 & 22 & 0.614 \\
5  & 0.576 & 14 & 0.584 & 23 & 0.627 \\
6  & 0.602 & 15 & 0.589 & 24 & 0.623 \\
7  & 0.627 & 16 & 0.577 & 25 & 0.612 \\
8  & 0.657 & 17 & 0.604 & 26 & 0.589 \\
9  & 0.667 & 18 & 0.568 & 27 & 0.581 \\
\bottomrule
\end{tabular}
\end{table}

\clearpage
\section{Budget curves on WikiText and arXiv}
\label{app:budget}

\begin{table}[H]
\centering
\caption{WikiText budget curve (Qwen2.5-1.5B, $\alpha$ from WikiText
calibration). NoHairCache has the lowest PPL at every budget; the margin over
StreamingLLM narrows toward full cache.}
\label{tab:budget-wikitext}
\begin{tabular}{S[table-format=4.0] S[table-format=2.1] S[table-format=2.3] S[table-format=2.3] S[table-format=2.3]}
\toprule
{Budget} & {Retained (\%)} & {H2O} & {NoHairCache} & {StreamingLLM} \\
\midrule
32 & 1.5 & 14.853 & \bfseries 12.695 & 13.041 \\
64 & 3.1 & 13.910 & \bfseries 11.602 & 12.056 \\
128 & 6.2 & 12.799 & \bfseries 10.830 & 11.073 \\
256 & 12.5 & 11.726 & \bfseries 10.200 & 10.289 \\
512 & 25.0 & 10.764 & \bfseries 9.691 & 9.748 \\
1024 & 50.0 & 9.874 & \bfseries 9.369 & 9.373 \\
\bottomrule
\end{tabular}
\end{table}

\begin{table}[H]
\centering
\caption{arXiv budget curve (Qwen2.5-1.5B, $\alpha$ from WikiText
calibration). StreamingLLM wins only at the tightest budget (32),
NoHairCache wins at all larger budgets.}
\label{tab:budget-arxiv}
\begin{tabular}{S[table-format=4.0] S[table-format=2.1] S[table-format=2.3] S[table-format=2.3] S[table-format=2.3]}
\toprule
{Budget} & {Retained (\%)} & {H2O} & {NoHairCache} & {StreamingLLM} \\
\midrule
32 & 1.5 & 7.388 & 6.298 & \bfseries 6.247 \\
64 & 3.1 & 6.879 & \bfseries 5.476 & 5.605 \\
128 & 6.2 & 6.228 & \bfseries 4.855 & 4.978 \\
256 & 12.5 & 5.502 & \bfseries 4.387 & 4.525 \\
512 & 25.0 & 4.786 & \bfseries 4.066 & 4.170 \\
1024 & 50.0 & 4.196 & \bfseries 3.911 & 3.948 \\
\bottomrule
\end{tabular}
\end{table}

\section{Proposition 2 proof (full)}
\label{app:prop2}

This appendix complements the condensed proof of Proposition~\ref{prop:sliding_window} in Sec.~\ref{sec:score-family} with the complete derivation and an explicit statement of its single-step limit. The multi-step trajectory extension is not used and the single-step limit is used only for the endpoint interpretation in Sec.~\ref{sec:score-family}.

\noindent\textbf{Assumption.} At step $t$, for all $i\neq j$,
$A_i/A_j\in(0,\infty)$.

\noindent\textbf{Proof.} Fix $i<j$; then $t-i>t-j$, so
$\rho_{ij}:=(t-j)/(t-i)\in(0,1)$.
The score ratio is
\begin{equation}
r_{ij}(\alpha)=\frac{s_i}{s_j}=\frac{A_i}{A_j}\,\rho_{ij}^{\alpha}.
\end{equation}
Since $\rho_{ij}^{\alpha}\to0$ as $\alpha\to\infty$ and $A_i/A_j$ is a
positive constant, $r_{ij}(\alpha)\to0$.
Moreover, $r_{ij}(\alpha)<1$ iff
$\rho_{ij}^{\alpha}<A_j/A_i$ iff
$\alpha\ln\rho_{ij}<\ln(A_j/A_i)$; since $\ln\rho_{ij}<0$, this is
$\alpha>\alpha_{ij}$ with
$\alpha_{ij}=\ln(A_i/A_j)/\ln(1/\rho_{ij})$.
Let $\alpha^*=\max_{i<j}\alpha_{ij}<\infty$ (finitely many pairs).
For all $\alpha>\alpha^*$, every $i<j$ satisfies $s_i<s_j$, so the score is
strictly decreasing in age and the top-$k$ set is $\{t-k+1,\dots,t\}$.
\hfill$\square$

\noindent\textbf{Remark.} This is a single-step limit, which we use only for the endpoint
interpretation in Sec.~\ref{sec:score-family}; multi-step trajectories require
an additional fixed-point argument, which we leave to future work and do not
use in the main text.
\section{LongBench four-way multiple-choice details}
\label{app:longbench}

Distractors are sampled from other questions' ground-truth answers
(3 per question, excluding the correct answer) and options are shuffled.
Accuracy is computed by comparing the log-probability of each option
after the prompt ``Answer:''.
Qasper is included for completeness but sits near the Qwen2.5-1.5B
capability floor (full-cache accuracy $0.49$), and does not discriminate
between eviction policies.

\bibliographystyle{plainnat}
\bibliography{refs}

@article{li2024instability,
  title={Measuring and Controlling Instruction (In)Stability in Language Model Dialogs},
  author={Li, Kenneth and Liu, Tianle and Bashkansky, Naomi and Bau, David and Vi{\'e}gas, Fernanda and Pfister, Hanspeter and Wattenberg, Martin},
  journal={arXiv preprint arXiv:2402.10962},
  year={2024}
}

@article{liu2024lost,
  title={Lost in the Middle: How Language Models Use Long Contexts},
  author={Liu, Nelson F and Lin, Kevin and Hewitt, John and Paranjape, Ashwin and Bevilacqua, Michele and Petroni, Fabio and Liang, Percy},
  journal={Transactions of the Association for Computational Linguistics},
  volume={12},
  pages={157--173},
  year={2024}
}

@inproceedings{xiao2024streaming,
  title={Efficient Streaming Language Models with Attention Sinks},
  author={Xiao, Guangxuan and Tian, Yuandong and Chen, Beidi and Han, Song and Lewis, Mike},
  booktitle={International Conference on Learning Representations (ICLR)},
  year={2024}
}

@article{barbero2025first,
  title={Why Do {LLMs} Attend to the First Token?},
  author={Barbero, Federico and Arroyo, Alvaro and Gu, Xiangming and Perivolaropoulos, Christos and Bronstein, Michael and Veli{\v{c}}kovi{\'c}, Petar and Pascanu, Razvan},
  journal={arXiv preprint arXiv:2504.02732},
  year={2025}
}

@inproceedings{kobayashi2020norm,
  title={Attention is Not Only a Weight: Analyzing Transformers with Vector Norms},
  author={Kobayashi, Goro and Kuribayashi, Tatsuki and Yokoi, Sho and Inui, Kentaro},
  booktitle={Proceedings of the 2020 Conference on Empirical Methods in Natural Language Processing (EMNLP)},
  pages={7057--7075},
  year={2020}
}

@inproceedings{abnar2020flow,
  title={Quantifying Attention Flow in Transformers},
  author={Abnar, Samira and Zuidema, Willem},
  booktitle={Proceedings of the 58th Annual Meeting of the Association for Computational Linguistics (ACL)},
  pages={4190--4197},
  year={2020}
}

@inproceedings{dong2021rank,
  title={Attention is Not All You Need: Pure Attention Loses Rank Doubly Exponentially with Depth},
  author={Dong, Yihe and Cordonnier, Jean-Baptiste and Loukas, Andreas},
  booktitle={International Conference on Machine Learning (ICML)},
  pages={2793--2803},
  year={2021}
}

@inproceedings{noci2022signal,
  title={Signal Propagation in Transformers: Theoretical Perspectives and the Role of Rank Collapse},
  author={Noci, Lorenzo and Anagnostidis, Sotiris and Biggio, Luca and Orvieto, Antonio and Singh, Sidak Pal and Lucchi, Aurelien},
  booktitle={Advances in Neural Information Processing Systems (NeurIPS)},
  volume={35},
  pages={27198--27211},
  year={2022}
}

@inproceedings{barbero2024glasses,
  title={Transformers Need Glasses! Information Over-Squashing in Language Tasks},
  author={Barbero, Federico and Banino, Andrea and Kapturowski, Steven and Kumaran, Dharshan and Ara{\'u}jo, Jo{\~a}o GM and Vitvitskyi, Alex and Pascanu, Razvan and Veli{\v{c}}kovi{\'c}, Petar},
  booktitle={Advances in Neural Information Processing Systems (NeurIPS)},
  year={2024}
}

@article{nakanishi2025ssmax,
  title={Scalable-Softmax Is a Superior Alternative to Softmax},
  author={Nakanishi, Ken M},
  journal={arXiv preprint arXiv:2501.19399},
  year={2025}
}

@article{lin2017critical,
  title={Critical Behavior from Deep Dynamics and Hidden Variables in Natural Language},
  author={Lin, Henry W and Tegmark, Max},
  journal={Entropy},
  volume={19},
  number={7},
  pages={299},
  year={2017}
}

@inproceedings{zhang2023h2o,
  title={{H2O}: Heavy-Hitter Oracle for Efficient Generative Inference of Large Language Models},
  author={Zhang, Zhenyu and Sheng, Ying and Zhou, Tianyi and Chen, Tianlong and Zheng, Lianmin and Cai, Ruisi and Song, Zhao and Tian, Yuandong and R{\'e}, Christopher and Barrett, Clark and Wang, Zhangyang and Chen, Beidi},
  booktitle={Advances in Neural Information Processing Systems (NeurIPS)},
  volume={36},
  year={2023}
}

@inproceedings{liu2023scissorhands,
  title={Scissorhands: Exploiting the Persistence of Importance Hypothesis for {LLM} {KV} Cache Compression at Test Time},
  author={Liu, Zichang and Desai, Aditya and Liao, Fangshuo and Wang, Weitao and Xie, Victor and Xu, Zhaozhuo and Kyrillidis, Anastasios and Shrivastava, Anshumali},
  booktitle={Advances in Neural Information Processing Systems (NeurIPS)},
  volume={36},
  year={2023}
}

@inproceedings{li2024snapkv,
  title={{SnapKV}: {LLM} Knows What You are Looking for Before Generation},
  author={Li, Yuhong and Huang, Yingbing and Yang, Bowen and Venkitesh, Bharat and Locatelli, Acyr and Ye, Hanchen and Cai, Tianle and Lewis, Patrick and Chen, Deming},
  booktitle={Advances in Neural Information Processing Systems (NeurIPS)},
  year={2024}
}

@article{zhang2024pyramidkv,
  title={{PyramidKV}: Dynamic {KV} Cache Compression based on Pyramidal Information Funneling},
  author={Zhang, Yichi and Gao, Bofei and Liu, Tianyu and Lu, Keming and Xiong, Wayne and Dong, Yue and Chang, Baobao and Hu, Junjie and Xiao, Wen and others},
  journal={arXiv preprint arXiv:2406.02069},
  year={2024}
}

@inproceedings{ge2024model,
  title={Model Tells You What to Discard: Adaptive {KV} Cache Compression for {LLMs}},
  author={Ge, Suyu and Zhang, Yunan and Liu, Liyuan and Zhang, Minjia and Han, Jiawei and Gao, Jianfeng},
  booktitle={International Conference on Learning Representations (ICLR)},
  year={2024}
}

@article{xiang2026nohair,
  title         = {A Conditional No-Hair Theorem for Transformers: How Language Models Forget Their Prompt},
  author        = {Xiang, Shuyang},
  year          = {2026},
  journal       = {arXiv preprint arXiv:2610.24133},
  eprint        = {2610.24133},
  archivePrefix = {arXiv},
  primaryClass  = {cs.CL}
}

\end{document}